# Multi-strip observation scheduling problem for active-imaging agile earth observation satellites

Zhongxiang Chang[1,2,3], Abraham P. Punnen[2*], Zhongbao Zhou[1,3]

[1]School of Business Administration, Hunan University, Changsha, China, 410082

[2]Department of Mathematics, Simon Fraser University, Surrey, BC, Canada, V3T 0A3

[3]Hunan Key Laboratory of intelligent decision-making technology for emergency management, Changsha, China, 410082

## Abstract

Active-imaging agile earth observation satellite (AI-AEOS) is a new generation agile earth observation satellite (AEOS). With renewed capabilities in observation and active imaging, AI-AEOS improves upon the observation capabilities of AEOS and provide additional ways to observe ground targets. This however makes the observation scheduling problem for these agile earth observation satellite more complex, especially when considering multi-strip ground targets. In this paper, we investigate the multi-strip observation scheduling problem for an active-image agile earth observation satellite (MOSP). A bi-objective optimization model is presented for MOSP along with an adaptive bi-objective memetic algorithm which integrates the combined power of an adaptive large neighborhood search algorithm (ALNS) and a nondominated sorting genetic algorithm II (NSGA-II). Results of extensive computational experiments are presented which disclose that ALNS and NSGA-II when worked in unison produced superior outcomes. Our model is more versatile than existing models and provide enhanced capabilities in applied problem solving.

---

[*]Corresponding author: Abraham P. Punnen, email: apunnen@sfu.ca

# 1. Introduction

The active-imaging agile earth observation satellite (AI-AEOS) is a state-of-the-art agile earth observation satellite (AEOS), having three-axis (roll, pitch, yaw) attitude maneuvering capability (Lemaître, Verfaillie, Jouhaud, Lachiver, & Bataille, 2002) and active imaging capabilities (Chang, Chen, Yang, & Zhou, 2020; Yang, Chen, He, Chang, & Chen, 2018). Further, this class of satellites can modify their attitude angles (pitch, roll and yaw) during observation. Active imaging is distinguished from the traditional passive imaging, which is required to hold the attitude angles during the whole observation tenure (L He, Liu, Laporte, Chen, & Chen, 2018; X. L. Liu, Laporte, Chen, & He, 2017; Peng et al., 2019; P. Wang, Reinelt, Gao, & Tan, 2011). Because of the active imaging capability, AI-AEOS can observe a ground target along arbitrary directions (direction of observation strips (DOS)) during its visible time window (VTW) (see Figure 1(a)). For an AI-AEOS, the number of DOS are infinite and could take values from [0°, 360°]. The observation of AI-AEOS is referred to as *non-along track observation* (NATO). AEOS without active imaging capabilities (See Figure 1(b)) observes ground targets on its flight passively, and its DOS must be parallel to its track, and its observations are categorized as *along track observation* (ATO).

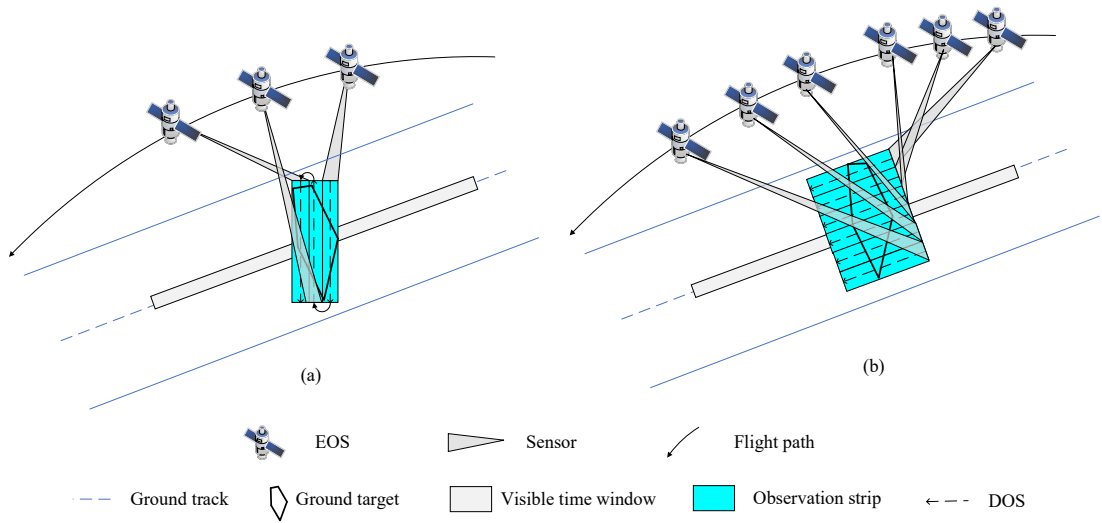

Figure 1 Non-along track observation and Along track observation

The number of observation strips for covering the ground target by AI-AEOS (3 observation strips) is significantly less than that of AEOS which does not have the capability of active imaging (6 observation strips) (See Figure 1), which may be beneficial to observe more ground targets. Moreover, AI-AEOS needs to modify its attitude angles during observation of each strip to realize NATO and therefore NATO will spend more energy than ATO (D. Wu et al., 2020). NATO provides additional observation prospects (Yang et al., 2018) for imaging ground targets, and even some of these prospects may in fact spend less energy for some of the observation strips. However, the number of observation strips decreases, and the observation attitudes could worsen, (See Figure 1). Worse observation attitudes may require more time to maneuver, which in turn will cost more energy and likely to decrease the image quality. Thus, any useful mathematical model for the problem needs to take this into consideration. Because of the infinite number of DOS, NATO makes observation scheduling problem for an active-imaging agile earth observation satellite (OSPFAIAS) more complex than the normal OSPFAS, which itself is known to be NP-Hard (Lemaître et al., 2002).

In (Yang et al., 2018), a simplified version of OSPFAIAS has been studied that considers two specific DOS for observing the ground target areas. To the best of our knowledge, there are no other works on OSPFAIAS that utilizes NATO capabilities effectively in observing more ground targets (or gaining higher image quality) with less energy. The primary goal of this paper is to reduce this gap and develop a practical and effective heuristic algorithm for OSPFAIAS.

Following this goal and the line of research, we explore OSPFAIAS with the objectives of improving the cumulative image quality and energy consumption. In particular, we will focus on multi-strip ground targets which need to be covered by several observation strips. Thus, we call OSPFAIAS as *the multi-strip observation scheduling problem for an active-imaging agile earth observation satellite* (MOSP).

The major contributions of this paper are summarized as below.

(1) A novel evaluation function is presented to calculate the cumulative image quality of AI-AEOS.
(2) Two types of energy consumption measures during observation of AI-AEOS are considered and an evaluation function is proposed to calculate the combined energy consumption.
(3) Four types of partition sets, namely ATO partition, NATO partition, Complete partition, and Envelope partition, are considered and analyze the efficacy of each one of them in detail.
(4) To optimize the cumulative image quality and overall energy consumption simultaneously, we model MOSP as a bi-objective optimization problem.

(5) Exploiting the special characteristics of MOSP, we propose an adaptive bi-objective memetic algorithm, ALNS+NSGA-II, to solve it.

(6) Extensive computational results are presented using our algorithm which establish the superiority of the proposed algorithm in obtaining good quality schedules under reasonable running time.

The rest of the paper is organized as follows. In Section 2 we present a brief review of related literature. In section 3, we develop two objective functions that absorb our optimization goals and then a bi-objective optimization model is developed to represent MOSP. Section 4 deals with the design of our adaptive multi-objective memetic algorithm, ALNS+NSGA-II which uses the operators, "Destroy" and "Repair", to generate adaptive large neighborhoods for evolution. Section 5 reports results of systematic experimental analysis carried out using our algorithm followed by concluding remarks in Section 6.

## 2. Literature Review

To the best of our knowledge, no publications exist in literature that consider capabilities of the AI-AEOS in the generalized form as presented in this paper. However, several special cases of the problem with simplifying assumptions are available. While such assumptions were valid 10 to 20 years ago, observation scheduling problem for modern-day satellites requires to consider a different set of capabilities and objectives. Wolfe and Sorensen (Wolfe & Sorensen, 2000) is the first published work that focussed on the OSPFAIAS . They considered spot/small area ground targets, which can be coved by an AI-AEOS in one observation strip. The imaging duration for each ground target belongs to the interval $[d_{min}, d_{max}]$, where $d_{min}$ and $d_{max}$ respectively denotes minimum and maximum image duration. They proposed a trapezoidal suitability method, by considering the benefits of different observation windows (OWs) during the *visible time window* (VTW). They also proposed three types of algorithms, a priority dispatch algorithm (PD), a look ahead algorithm (LA), and a genetic algorithm (GA), to solve the problem. The problems addressed in (Wolfe & Sorensen, 2000) excluded the transition time between different observations and the energy consumption. Although such an approach was appropriate in the year 2000, for current mission planning problems using modern day satellites, ground targets are always more than what the existing in-orbit EOSs can observe (X. W. Wang, Chen, & Han, 2016). Moreover, the transition time between different observations is time-dependent (Chang, Zhou, Xing, & Yao, 2021; X. L. Liu et al., 2017; Lu, Chang, Zhou, & Yao, 2021) and on-board power is not sufficient for infinite observations (Zhao, Wang, Hao, & Wang, 2019). Our study in this paper takes into consideration all these capabilities and constraints.

Bunkheila et al. (Bunkheila, Ortore, & Circi, 2016) analyzed the mission planning problem of AEOS in detail. They partitioned the problem into two sub-problems: geometric analysis to strip the ground targets and temporal analysis to calculate the visible time windows. Although AEOS studied in (Bunkheila et al., 2016) can perform active imaging, their focus was primarily on acquisition operations and did not consider any constraint between different ground targets or observation tasks.

Cui, Xiang, and Zhang. (Cui, Xiang, & Zhang, 2018) studied the mission planning problem of video imaging satellites. They proposed a constraint satisfaction model using short-period planning (1/6 orbit) and solved the problem using an ant colony based search algorithm with a taboo list. However the transition time used in their experiments was set to a constant and was not time-dependent. Further, in their experiments, mostly simplified instances were considered with two fixed image durations for observing each ground target.

For active-imaging agile earth observation satellites, Nag et al. (Nag, Li, & Merrick, 2018) extended the image duration for each unique image in their simulation experiment. However, their model did not consider observation scheduling with the variable image durations for each unique image. In a follow up experimental study, Nag et al. (Nag et al., 2019) considered on board calculations while retaining the earlier model structure and algorithm from (Nag et al., 2018).

The AEOS(s) considered in most of the literature did not take into consideration active imaging capabilities, and the direction of observation strips (DOS) must be parallel to the track of AEOS(s). Considering the area of ground targets, the existing research works on the topic can be classified into two groups: small ground targets (points or small areas) that AEOS can be observed in one pass (Augenstein, Estanislao, Guere, & Blaes, 2016; Chang, Chen, Yang, & Zhou, 2019; Chang, Zhou, Li, Xiao, & Xing, 2022; Chang, Zhou, Yao, & Liu, 2021; Lei He, Liu, Chen, Xing, & Liu, 2019; S. K. Liu & Yang, 2019; Shao et al., 2018; J. J. Wang,

Demeulemeester, Hu, Qiu, & Liu, 2019; J. J. Wang, Zhu, Yang, Zhu, & Ma, 2015; K. Wu, Zhang, Chen, Chen, & Shao, 2019; R. Xu, Chen, Liang, & Wang, 2016), and large ground targets, which need to be decomposed to smaller pieces and then observed by AEOS(s) in several passes (Berger, Lo, & Barkaoui, 2020; Niu, Tang, & Wu, 2018; Y. J. Xu, Liu, He, & Chen, 2020). In this paper, we only consider small area ground targets, which can be coved by AEOS (with active imaging capabilities, i.e. AI-AEOS) in one pass, but the number of observation strips for covering them is unrestricted.

Some authors ignored the transition time (Tangpattanakul, Jozefowiez, & Lopez, 2015; Wolfe & Sorensen, 2000) of AEOS or assumed it as a constant (Chang et al., 2020; Hu, Zhu, An, Jin, & Xia, 2019; Lemaître et al., 2002; H. Wang, Yang, Zhou, & Li, 2019). Because of the flexible attitude maneuver of AEOS, more and more authors view the transition time of AEOS as a time-dependent variable (L He, de Weerdt, & Yorke-Smith, 2019; L He et al., 2018; Li & Li, 2019; X. L. Liu et al., 2017; Peng et al., 2019; Peng, Song, Xing, Gunawan, & Vansteenwegen, 2020; S. Wang, Zhao, Cheng, Zhou, & Wang, 2019). They proposed several methods for calculating the transition time but all of them are essentially equivalent. The equation proposed in (X. L. Liu et al., 2017) for this calculation is

$$trans(\Delta g) = \begin{cases} 35/3 & \Delta g \leq 10 \\ 5 + \frac{\Delta g}{v_1} & 10 < \Delta g \leq 30 \\ 10 + \frac{\Delta g}{v_2} & 30 < \Delta g \leq 60 \\ 16 + \frac{\Delta g}{v_3} & 60 < \Delta g \leq 90 \\ 22 + \frac{\Delta g}{v_4} & \Delta g > 90 \end{cases} \quad (1)$$

where the unit of transition time is in seconds. $v_1, v_2, v_3$ and $v_4$ are four different angular transition velocities and the values of them are $v_1 = 1.5°/s$, $v_2 = 2°/s$, $v_3 = 2.5°/s$ and $v_4 = 3°/s$. $\Delta g$ denotes the total change of attitude angles between two observations and is calculated by

$$\Delta g = \Delta \pi + \Delta \gamma + \Delta \psi \quad (2)$$

where $\Delta \pi$, $\Delta \gamma$, and $\Delta \psi$ indicate the change of the pitch angle, the roll angle, and the yaw angle, respectively. The attitude capability of AI-AEOS studied in our paper is much more versatile than that of AEOS(s) generally studied in literature, but the attitude maneuver is the same as earlier studies and ignore the angular transition velocities. It may be noted that there is no AEOS currently in-orbit having the capability of active imaging in China (the source of our data) and hence we did not calibrate the exact parameter of the angular transition velocities of AI-AEOS, instead we adopted equation (1) in our work.

## 3. Model Development

In this section we develop a bi-objective optimization model for MOSP. We first summarize some reasonable assumptions we made to standardize and simplify MOSP to make our work consistent with existing research in the area and follow current engineering practices.

(1) AI-AEOS can modify the attitude angles during observations, but we do not pay attention to the attitude maneuver during observations, which is the focus of attitude scheduling (D. Wu et al., 2020), and we assume that the attitude maneuver is feasible in the whole VTW.

(2) Since different types of sensors have different attributions, we only consider a single optical AI-AEOS in our study.

(3) AI-AEOS has sufficient on-board memory during the entire scheduling horizon and hence it is not necessary to consider satellite image data downlink.

(4) As mentioned earlier, we only consider small ground targets that AI-AEOS can be observed in one pass, while the number of observation strips for covering them is unrestricted. Since large ground targets need to be decomposed before observation scheduling (Berger et al., 2020; Niu et al., 2018; Y. J. Xu et al., 2020), we can discard large ground targets and our restriction to small ground targets is without loss of generality.

(5) Since ground targets are always larger than what the existing in-orbit EOSs can observe (X. W. Wang et al., 2016), we assume that all ground targets can be observed at most once.

(6) We assume that the sensor is always on during the whole scheduling horizon. This way, it is not necessary to spend extra time and energy for (re)starting the sensor before every observation.

(7) All ground targets are observed with uniform speed, no matter it is NATO or ATO, and the length of all observation strips for the same ground target is the same.

Based on the aforementioned assumptions, the observation scheme $S$ for the MOSP can be described by the set:

$$\mathcal{S} = \{Sat, St, Et, AGT\} \tag{3}$$

and the parameters used in equation (3) are discussed below.

- $Sat = \{sat, |S| = n_s\}$ is the set of earth observation satellites (EOSs) for observation scheduling. Each $sat$ is represented as a three-element set

$$sat = \{Id, type, A\} \tag{4}$$

where $Id$ is the identifier of $sat$, $type$ represents the type of sensor installed on $sat$, and by assumption (1), $type = optical$, and finally, $A$ represents the basic attributes of the sensor installed in EOS. Different types of the sensor have different attributes (Lei He et al., 2019; X. L. Liu et al., 2017; P. Wang et al., 2011). The collection of attributes $A$ of an optical sensor is composed of five elements which is represented as:

$$A = \{\theta, \gamma, \pi, \psi, d_0\} \tag{5}$$

where $\theta$ denotes the cone angle of the optical camera, which is related to the field of view and $\gamma$, $\pi$ and $\psi$ indicates the maximum angle of roll, pitch and yaw respectively. These three angles restrict the visibility of EOS with ground targets. Finally, $d_0$ is the minimum image duration for each observation strip.

- Let $[St, Et]$ be the scheduling horizon for the MOSP. In our work, the scheduling horizon is from 2020/08/01 00:00:00 to 2020/08/02 00:00:00.

- The set $AGT = \{gt_i | 1 \leq i \leq n_{gt}, |AGT| = n_{gt}\}$ indicates the collection of all ground targets considered for the scheduling horizon and $n_{gt}$ represents the total number of ground targets considered. Each ground target $gt_i$ is represented by

$$gt_i = \{Id, \omega, s, e, wc, ow\} \tag{6}$$

where $Id$ is the identifier of the ground target $gt_i$, $\omega$ reflects the priority of $gt_i$, and $s$ and $e$ represents begin time and end time of the visible time window (VTW) of $gt_i$ respectively. Further, $wc$ is the original workpiece congestion of $gt_i$ as defined in the function (29), and $ow$ reflects the observation window (OW) for observing $gt_i$ defined by the function (7) and the time span of $ow$ belongs to the corresponding VTW ($[s, e]$). The parameter $ow$ depends on various quantities and is represented by

$$ow = \{Id, way, dos, b, e, \pi_o, \gamma_o, \psi_o, \pi_\infty, \gamma_\infty, \psi_\infty, osList\} \tag{7}$$

where $Id$ is an identifier of $ow$, which is used to distinguish several OWs[1] belonging to the same ground target and $way$ is a binary variable that represents the observation ways of $ow$. $way = 0$ implies ground target will be observed passively, otherwise it will be observed actively. $dos$ indicates the direction of observation strips (DOS), and as mentioned earlier, $dos$ belongs to the interval [0°,360°]. Note that, when $way = 0$, the value of $dos$ is invalid. $b$ and $e$ respectively denotes the begin moment and the end moment of $ow$. $\pi_o$, $\gamma_o$, $\psi_o$, $\pi_\infty$, $\gamma_\infty$ and $\psi_\infty$ indicates the begin pitch angle, begin roll angle, begin yaw angle, end pitch angle, end roll angle and end yaw angle of $ow$ respectively. $osList$ is a set of observation strips for covering the ground target $gt_i$ under DOS equals $dos$. By assumption (4), all observation strips must be observed completely one by one. In addition, all observation strips are stored in a consistent sequence, left-to-right or right-to-left, which is up to the distance of the left/right observation strip with the sub-track of EOS. Every observation strip is composed of ten elements and hence $os$ is represented by

$$os = \{Id, d, sp, ep, \pi_o, \gamma_o, \psi_o, \pi_\infty, \gamma_\infty, \psi_\infty\} \tag{8}$$

where $Id$ is the identifier of $os$ and $d$ denotes the image duration of $os$. Note that, as mentioned in assumption (7), the image duration of all observation strips in the same $osList$ must be the same. $sp$ and $ep$ respectively denotes the start center point and the end center point of $os$, and these values are used to calculate the real-time attitude angles. $\pi_o$, $\gamma_o$, $\psi_o$, $\pi_\infty$, $\gamma_\infty$ and $\psi_\infty$ indicates the begin pitch angle, begin roll angle, begin yaw angle, end pitch angle, end roll angle and end yaw angle of $os$ respectively after confirming observation moment.

## 3.1. Cumulative image quality

Without the capability of active imaging, the quality of the image acquired mainly depended on the observation begin moment (L He et al., 2019; L He et al., 2018; X. L. Liu et al., 2017; Peng et al., 2019; Peng et al., 2020), because AEOS has to maintain the values of the attitude angles during the entire OW. The image quality ($q$) belongs to the interval [0,1]. Peng et al. (Peng et al., 2020) (see also (X. L. Liu et al., 2017) and (L He et al., 2018)) presented an *image quality function*, which depends on the observation pitch angle, is given by

$$q(u) = 1 - \frac{|\pi(u)|}{90} \tag{9}$$

where $\pi(u)$ denotes the pitch angle when AEOS observes the ground target at the observation begin moment $u$.

---
[1] A ground target with several VTWs will be clone as several ground targets with the same identifier but different VTWs.

As mentioned above, the image quality in the function (9) depends only on the begin pitch angle, while AI-AEOS can modify the attitude angles during the whole OW. Therefore, the image quality for AI-AEOS should be redefined as a cumulative variable and is related to the pitch angles and roll angles during the whole OW yielding

$$Q(ow) = \frac{\sum_{u \in ow} q(u)}{MIQ} \tag{10}$$

where $ow$ denotes the observation window (OW), $q(u)$ represents the instant image quality obtained at the moment $u$, which is redefined as the function (11)[2], $\sum_{u \in ow} q(u)$ indicates the cumulative image quality obtained by an observation, and $MIQ$ indicates an upper bound. By the way, $Q(ow)$ also belongs to the interval [0,1] and the bigger its value the better the quality of the image. The value $q(u)$ is defined as

$$q(u) = \left(1 - \frac{|\pi(u)|}{90}\right) \times \left(1 - \frac{|\gamma(u)|}{90}\right) \tag{11}$$

where $\pi(u)$ and $\gamma(u)$ denotes the instant pitch angle and the instant roll angle respectively when the AI-AEOS observes the ground target at moment $u$. In addition, the instant image quality $q(u)$ belongs to the interval [0,1] and as $q(u)$ increases the instant image quality improves. Note that when $ow.way = 0$, the value of $q(u)$ does not change with changes in $u$, and it is similar to the image quality defined in (L He et al., 2018; X. L. Liu et al., 2017; Peng et al., 2019). Note that the notation $ow.way$ means the value $way$ associated with the specific $ow$. For the rest of this paper, we use this type of notation, adapted from object-oriented programming, without further explanation. We can certainly use subscripts to achieve the same but preferred our notational convention for simplicity of presentation.

### 3.2. Overall energy consumption

By assumption (6), two types of energy consumptions are considered: energy consumption by observation (EO) and energy consumption by attitude conversion (EA). To facilitate the calculation of these two types of energy consumption values for an observation scheme $S$, some additional notations and terminologies are introduced below.

---

[2] According to (D. Wu et al., 2020), we know that the yaw angle does not reflect the image quality, so we do not consider it into the function (11).

- $E$     total energy consumption of $S$
- $ot_a$     total observation time of $S$ in active-imaging
- $ot_p$     total observation time of $S$ in passive-imaging
- $at_{in}$     total attitude conversion time between multiple strips in all ground targets
- $at_{out}$     total attitude conversion time between every two adjacent ground targets
- $eo_a$     energy consumption rate for observation in active-imaging
- $eo_p$     energy consumption rate for observation in passive-imaging
- $ea$     energy consumption rate for attitude conversion
- $GT$     the set of ground targets scheduled to be observed

The total energy consumption of the observation scheme $S$ is the sum of EO and EA and is defined by the function (12)

$$E = eo_a \times ot_a + eo_p \times ot_p + ea \times (at_{in} + at_{out}) \tag{12}$$

where $eo_a$, $eo_p$ and $ea$ are constants with $ea = 0.05W$. It is apparent that active-imaging will spend more energy than passive-imaging since active-imaging needs to modify attitude angles at the same time. We set $eo_p = 0.08W$ and $eo_a = 0.1W$. Also, $ot_p$, $ot_a$, $at_{in}$ and $at_{out}$ are calculated respectively as follows.

$$\begin{cases} ot_p = \sum_{i=1}^{|GT|} \sum_{os \in gt_i.ow.osList} os.d & gt_i.ow.way == 0 \\ ot_a = \sum_{i=1}^{|GT|} \sum_{os \in gt_i.ow.osList} os.d & gt_i.ow.way == 1 \end{cases} \tag{13}$$

where $\sum_{os \in gt_i.ow.osList} os.d$ indicates the image duration for observing the ground target $gt_i$. $|GT|$ denotes the number of ground targets scheduled to be observed. Let

$$\Delta g_{gt_i \to gt_{i+1}} = |\, gt_{i+1}.ow.\pi_o - gt_i.ow.\pi_\infty\,| + |\, gt_{i+1}.ow.\gamma_o - gt_i.ow.\gamma_\infty\,| +$$
$$|\, gt_{i+1}.ow.\psi_o - gt_i.ow.\psi_\infty\,| \tag{14}$$

Let $\forall gt_i, gt_{i+1} \in GT$, which are two adjacently scheduled ground targets in $S^3$. Then $\Delta g$ in the equation (1) can be calculated using the function (14). So, $at_{out}$ can be calculated using

$$at_{out} = \sum_{i=1}^{|GT|-1} trans(\Delta g_{gt_i \to gt_{i+1}}) \tag{15}$$

The parameter $at_{in}$ is related to the number of observation strips of each ground target, and for each $gt_i \in GT$, the total attitude conversion time ($\Delta T_{gt_i}$) in $gt_i$ is calculated as the function

$$\Delta T_{gt_i} = \sum_{i=1}^{|gt_i.ow.osList|-1} trans(\Delta g_{os_i \to os_{i+1}}) \tag{16}$$

---
[3] Notably, all scheduled ground targets are sorted by their observation begin moment ascending.

where $|gt_i.ow.osList|$ represents the total quantity of observation strips in $gt_i$, $trans(\Delta g_{os_i \to os_{i+1}})$ indicates the attitude conversion time of two adjacent observation strips in $gt_i$ and $\Delta g_{os_i \to os_{i+1}}$ can be calculated by the function (14). Thus, $at_{in}$ can be calculated using the equation (17).

$$at_{in} = \sum_{i=1}^{|GT|-1} \Delta T_{gt_i} \tag{17}$$

Using Figure 1, the definition of NATO and ATO, and the function (1), it is easy to see that active imaging will spend the less attitude conversion time than passive imaging for fewer observation strips. In other words, active imaging will consume less energy than passive imaging for changing attitude conversion toward the same ground target.

## 3.3. A bi-objective optimization model

Three main decision variables considered in this model are $x_i$, $gt_i.ow.way$ and $gt_i.ow.b$. Here $x_i$ is a binary variable and represents whether ground target $gt_i$ is scheduled to be observed or not. The decision variable $gt_i.ow.way$ is a binary variable which indicates the observation ways of $ow$ for observing $gt_i$. In addition, $gt_i.ow.b$ denotes the observation begin moment of $gt_i$ and belongs to its visible time window (VTW) ( $[gt_i.s, gt_i.e]$ ).

Observing as much ground targets as possible was the original intention of the traditional OSPFAS (Lemaître et al., 2002). Therefore, the minimum loss rate of image quality (LR) is considered as one of the optimization objectives in our work, which is similar to maximizing the sum of image quality of scheduled ground targets.

$f_1(S)$, *the loss rate of image quality* (LR), considers the priority and image quality of scheduled ground targets. It belongs to the interval [0,1] and calculated using the equation

$$f_1(S) = 1 - \frac{\sum_{i=1}^{n_t} x_i \times gt_i.\omega \times Q(gt_i.ow)}{\sum_{j=1}^{n_{gt}} gt_j.\omega} \tag{18}$$

Here $\sum_{j=1}^{n_{gt}} gt_j.\omega$ is a constant parameter, which is calculated under the assumption that all ground targets are observed with highest image quality ($Q = 1$) and $\sum_{i=1}^{n_t} x_i \times gt_i.\omega \times Q(gt_i.ow)$ denotes the sum of priority and image quality of scheduled ground targets in their OW.

Consumption of less energy is another goal for OSPFAS (J. J. Wang et al., 2015). Minimizing the energy consumption (EC) is not only good for observation, it is also good for

the AEOS system to operate smoothly. So we consider minimizing EC as another optimization objective. Let

$$f_2(S) = \frac{E}{MEC} \tag{19}$$

where $E$ denotes the total energy consumption to observe all scheduled ground targets as defined in the function (12). A constant parameter, named maximum energy consumption (*MEC*), is included to normalize $E$. MEC is calculated under the assumption that all ground targets are observed during their whole VTW. After the normalization, $f_2(S)$ also belongs to the interval [0,1]. The value of *MEC* is calculated by

$$MEC = n_{gt} \times eo_a \times \max_{gt_i \in AGT}(gt_i.e - gt_i.s) + maxT \times (1 + MaxOS) \times ec \times n_{gt} \tag{20}$$

where $maxT$ denotes the maximum transition time and is set as 100s according to the function (1). Here, $maxT \times ec \times n_{gt}$ indicates the attitude conversion time between all adjacent ground targets, while $maxT \times MaxOS \times ec \times n_{gt}$ represents the attitude conversion time between multiple strips of all ground targets, and $MaxOS = 10$ denotes the maximum number of observation strips. On the other hand, $\max_{gt_i \in AGT}(gt_i.e - gt_i.s)$ denotes the maximum length of VTW for all ground targets and $eo_a \times \max_{gt_i \in AGT}(gt_i.e - gt_i.s)$ indicates the maximum energy consumption of every ground target. Our objective is

$$min\ F(S) = \{f_1(S), f_2(S)\} \tag{21}$$

Note that we have two optimization objectives, *the loss rate of image quality* (LR) and *the energy consumption* (EC), as shown in the function (21)[4]. These two objectives are not irreconcilable in observation scheduling, so the simultaneous optimization of them is reasonable and possible. Note that, MOSP is a bi-objective discrete optimization problem (Kidd, Lusby, & Larsen, 2020). Let us now look at the constraints. First, we impose

$$gt_i.Id == gt_k.Id \Rightarrow x_i + x_k \leq 1 \qquad 0 \leq i, k \leq n_{gt} \tag{22}$$

The constraint (22) indicates that each ground target is observed at most once, which is consistent with assumption (5). The constraint

$$\begin{cases} gt_i.ow.b \geq gt_i.s \\ gt_i.ow.e \leq gt_i.e \end{cases} \quad 0 \leq i \leq n_{gt} \tag{23}$$

assures that all ground targets must be observed during their visible time window (VTW). Our

---

[4] Because of the normalization, the value of two objectives will be very small.

next constraint is

$$os_j.d \geq x_i \times sat.A.d_0 \quad os_j \in gt_i.ow.osList, 0 \leq i \leq n_{gt} \quad (24)$$

This indicates that if $x_i = 1$, the image duration of every observation strip for covering $gt_i$ cannot be smaller than the minimum image duration. Otherwise, this constraint is invalid. Combining (23) and (24) yields the complete constraint set to confirm observation time of each ground target. Our next constraint is

$$\begin{cases} x_j x_k \times (gt_k.ow.b - gt_j.ow.b) \leq x_j x_k & \forall k,j \in [0, n_{gt}] \\ x_j x_k \times (gt_j.ow.b - gt_k.ow.e) \geq x_j x_k \times trans\left(\Delta g_{gt_k \to gt_j}\right) \end{cases} \quad (25)$$

The inequalities in the constraint (25) represent the transition time constraint. The first inequality guarantees that the ground targets $gt_k$ and $gt_j$ are observed, and $gt_k$ is observed before $gt_j$ if and only if both of $x_j$ and $x_k$ are equal to 1. The second inequality represents the interval time between them must be bigger than the attitude conversion time from $gt_k$ to $gt_j$. If $x_k$ or $x_j$ equals zero, the constraint (25) will be redundant.

## 4. An adaptive bi-objective memetic algorithm

Memetic computing/algorithm (MC/MA) usually (Neri & Cotta, 2012) consists of an evolutionary framework and a set of local search algorithms, which are activated within the generation cycle. This algorithmic framework has been successfully used in solving various combinatorial optimization problems.

We design an adaptive bi-objective memetic algorithm, called ALNS+NSGA-II, to solve MOSP. The algorithm ALNS+NSGA-II combines an adaptive large neighborhood search algorithm (ALNS) and a nondominated sorting genetic algorithm II (NSGA-II). The basic framework of the algorithm is given as a flow chart in Figure 2.

ALNS is used as the local search structure within ALNS+NSGA-II to breed offspring solutions, because ALNS can search for good quality solutions faster and has been adopted in some of the earlier studies (L He et al., 2019; L He et al., 2018; Kadziński, Tervonen, Tomczyk, & Dekker, 2017; X. L. Liu et al., 2017; X. Wu & Che, 2019, 2020). The basic structure of ALNS is constructed around two loops. The inner loop is a local search process, which consists of various "Destroy" and "Repair" operators, while the outer loop uses some criteria to control the search process. A score and a weight is assigned to each operator, and an adaptive layer is included to update the weights and scores of different operators according to their performance in searching for better solutions. The "Destroy" operators and "Repair" operators are defined based on the characteristics of problem.

NSGA-II is adopted as the evolutionary mechanism for ALNS+NSGA-II to achieve Pareto frontier fast. Deb et al. (Deb, Pratap, Agarwal, & Meyarivan, 2002) improved NSGA (Srinivas & Deb, 1995) and proposed NSAG-II in 2002, which is one of the best known evolutionary multi-objective optimization algorithms (Gong, Jiao, Yang, & Ma, 2009). NSGA-II can obtain the Pareto frontier faster for a fast nondominated sorting approach, and they also proposed a fast-crowded distance estimation and a simple crowded comparison operator to address the shortcoming that the evolutionary algorithms are depended on parameters. Furthermore, NSGA-II has been adopted for several evolutionary computing paradigms(Kadziński et al., 2017; X. Wu & Che, 2019, 2020).

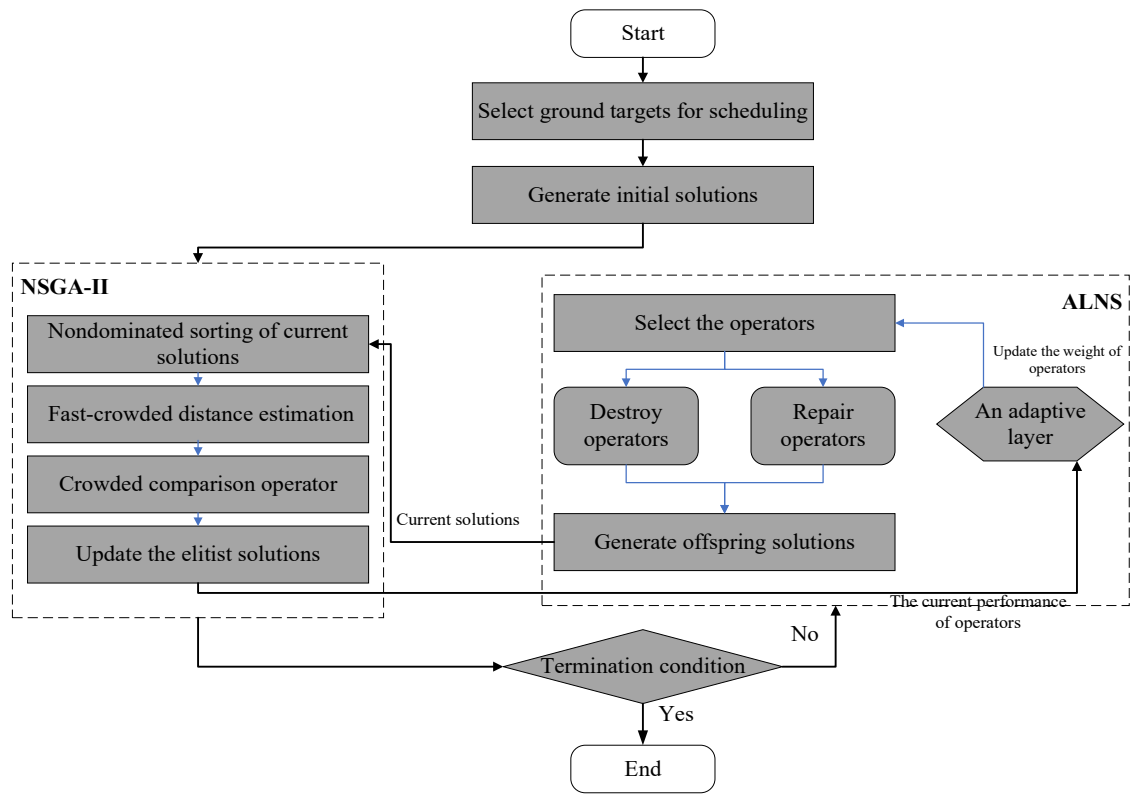

Figure 2 The basic flow of ALNS+NSGA-II

The pseudo code of ALNS+NSGA-II is shown as the Algorithm 1.

| Algorithm 1: ALNS+NSGA-II |
| --- |
| **Input:** A set of ground targets $AGT$, the probability of selecting ground target ($RS$), the population size of the initial solutions ($NS$), the archive solutions ($NA$) and the maximum quantity of iterations |

| | (*MaxIter*) |
|---|---|
| | **Output:** The elitist solutions $\Re s$ |
| 1: | **Repeat -------Generate initial solutions** |
| 2: | Select a set of ground targets (*IGT*) for observation scheduling from *AGT* randomly according to *RS*. |
| 3: | Adopting RGHA to generate a solution $S$ based on *IGT* |
| 4: | Justify whether the solution $S$ is existent and add the non-existent solution into $Ss$ |
| 5: | **End** until Size of ($Ss$) = NS+NA |
| 6: | Select and preserve the elitist solutions $\Re s$ according to the selection mechanism of NSGA-II |
| 7: | **Evolution-----Improve the solutions** |
| 8: | Select "Destroy" operators and "Repair" operators according to the adaptive layer |
| 9: | Based on the current elitist solutions, adopt selected operators to breed offspring solutions, the population of which is NS+NA |
| 10: | Combine the offspring solutions and the current elitist solutions |
| 11: | Update the elitist solutions $\Re s$ and the adaptive layer |
| 12: | **End** until iteration achieves $MaxIter$ |
| 13: | **Output** the latest elitist solutions $\Re s$ |

Our ALNS+NSGA-II adopts the fast-crowded distance estimation and the simple crowded comparison operator to select the elitist solutions (the line 6: and line 11: in Algorithm 1). All offspring solutions are generated (the line 9: in Algorithm 1) in the non-dominated space according to the box-method mentioned (to be discussed inin the section 4.4.).

We will now describe some of the crucial calculations in the algorithm such as the mechanism for selecting ground targets (line 2: in Algorithm 1), the initialization algorithm (line 3: in Algorithm 1), operators for breeding offspring solutions (line 9: in Algorithm 1), a selection mechanism for preserving offspring solutions line 6: and line 11: in Algorithm 1) and an adaptive layer line 8: and 11: in Algorithm 1) for controlling operators evolution.

## 4.1. Selecting ground targets

Since total number of ground targets are always larger than what the existing in-orbit EOSs can observe (X. W. Wang et al., 2016), we do not consider all ground targets (*AGT*) during every evolution cycle of observation scheduling and select a part of ground targets (*IGT*) from *AGT* randomly for scheduling. We let $RS \in [0,1)$ to control the selection process. A ground target is selected and added into IGT if and only if the random probability is larger than $RS$.

## 4.2. A randomized greedy heuristic algorithm (RGHA)

Since ALNS is not highly sensitive to the initial solution (Ropke, 2007) and the solution obtained by heuristic greedy algorithms is always feasible and stable, we propose a randomized greedy heuristic algorithm (RGHA) for constructing an initial solution. The pseudocode of RGHA is given in Algorithm 2.

Note that (line 4: in the Algorithm 2), the ATO strip partition method, proposed in (Xiao-dong, Ying-wu, Ren-jie, & Ju-fang, 2011), is adopted to generate ATO observation strips, while the NATO strip partition method of (Wen-yuan, Ren-jie, Xi-ying-zhi, & Hong-tao, 2016) is used to generate NATO observation strips[5]. To analyze the effect of the partition set proposed in (Wen-yuan et al., 2016), called *Envelop partition*, we design three other partition sets. The first one only considers ATO observation strips, named as *ATO partition*. The second one considers all NATO observation strips, named as *NATO partition*, and $dos$ of each ground target is set to the interval [0°, 360°] with steps of 30° each. The last one will consider all observation strips including ATO observation strips and NATO observation strips and called the *Complete partition*. We compared the efficiency of ALNS+NSGA-II under these partitions by extensive computational experiments and the results will be discussed in the next section. The analysis disclosed that the *Envelope partition* is more reasonable and suitable for solving MOSP.

| Algorithm 2 : A random greedy heuristic algorithm (RGHA) |
|---|
| **Input:** A set of ground targets $IGT$ |
| **Output:** An initial solution $S$, the element of which is a set of scheduled ground targets $GT$ |
| 1:   Sort all ground targets in $IGT$ by their priority descending. |
| 2:   **Repeat ------ Confirm observation moment of each ground target** |
| 3:       Choose a ground target $gt$ from $IGT$ one by one. |
| 4:       Get the observation strips from the corresponding partition set randomly. |
| 5:       Generate the observation moment of $gt$ randomly according to the constraints (23)-(24), and calculate the observation moment of all observation strips of $gt$. |
| 6:       Estimate whether $gt$ can be observed according to constraint (25), if the result is false, abandon $gt$ directly. Otherwise add $gt$ into $GT$. |
| 7:   **Until** all ground targets in $IGT$ are visited. |
| 8:   Calculate the value of two optimization objectives. |
| 9:   **Return** the initial solution $S$. |

---

[5] In addition, we do not have source code of them, but the DLL version of them.

## 4.3. "Destroy" and "Repair" operators

"Destroy" and "Repair" operators are used to generate adaptive large neighborhoods (ALNs) for optimization. "Destroy" operators change the elements (ground targets) of ALNs by deleting some observed ground targets, while "Repair" operators optimize the rank of all elements in ALNs and then repair parent solutions to generate offspring solutions. A variation of the box-method (Hamacher, Pedersen, & Ruzika, 2007), to be discussed in section 4.4, is adopted to decide whether new solutions are accepted or not. Considering the performance of every operator, an adaptive layer is designed in section 4.5 to control the utilization of the operators.

### 4.3.1. "Destroy" operators

In our implementation, four "Destroy" operators, **R-Destroy**, **Q-Destroy**, **E-Destroy** and **C-Destroy**, are defined according to different information that guide the algorithm. All deleted ground targets are saved into a set called taboo bank $B$ with a given size $|B|$, in which all ground targets will not be considered when we execute "Repair" operators. $B$ is empty before removing ground targets and filling $B$ to its full capacity is set as the termination criterion for "Destroy" operators. All unscheduled ground targets are saved in a set $F$. The ground targets in $F$ but not in $B$ will be selected by "Repair" operators to insert into a given solution and produce offspring solution.

**R-Destroy:** This operator deletes some scheduled ground targets from the given solution randomly.

**Q-Destroy:** The guiding information for this operator is the image quality of every scheduled ground target, which is calculated using (10). This operator ranks all scheduled ground targets by their image quality in ascending order and then deletes them one by one, from those representing the scheduled ground targets with lower image quality from the given solution in priority.

**E-Destroy:** The guiding information of this operator considers the observation energy consumption and original attitude conversion energy consumption and is named as $GI\_E$. For the ground target $gt_i$, $GI\_E$ can be calculated using

$$GI\_E(gt_i) = eo(gt_i) + ea(gt_i) \tag{26}$$

where $eo(gt_i)$ and $ea(gt_i)$ represents the observation energy consumption and original attitude conversion energy consumption of $gt_i$ respectively. According to (17), $eo(gt_i)$ can be calculated as.

$$eo(gt_i) = \begin{cases} eo_a \times \sum_{os \in gt_i.ow.osList} os.d & gt_i.ow.way == 1 \\ eo_p \times \sum_{os \in gt_i.ow.osList} os.d & gt_i.ow.way == 0 \end{cases} \quad (27)$$

and from (15), $ea(gt_i)$ can be defined as

$$ea(gt_i) = ea \times \left(trans(\Delta g_{o \to gt_i}) + trans(\Delta g_{gt_i \to o})\right) \quad (28)$$

where $ea$ denotes the energy consumption rate for attitude conversion, $o = \{0,0,0\}$ represents the original attitude status of AI-AEOS where the three attitude angles (pitch, roll, and yaw) are all zero. $trans(\Delta g_{o \to gt_i})$ represents the transition time spent by changing attitude angles from the original attitude status to the begin attitude angles of $gt_i$, while $trans(\Delta g_{gt_i \to o})$ indicates the transition time consumed by changing attitude angles from the end attitude angles of $gt_i$ to the original attitude status.

This operator ranks all selected ground targets by $GI\_E$ in descending order of energy consumption, and the ground target consuming more energy will be removed from the given solution in priority.

**C-Destroy:** The guiding information of this operator uses workpiece congestion defined in (Chang et al., 2020). For the ground target $gt_i$ the workpiece congestion ($gt_i.wc$) is defined as

$$gt_i.wc = \sum_{j=1}^{n} NoD(gt_j.\omega \times d_{ij}) \quad (29)$$

where $d_{ij}$ denotes the conflict distance between $gt_i$ and $gt_j$. If there is an irreconcilable conflict between them, i.e. either-or conflict, $d_{ij} = 1$; if there is a reconcilable conflict, $d_{ij} = 0.5$; else if there is no conflict, $d_{ij} = 0$. Note that $NoD()$ represents dimensionless processing defined as the function

$$NoD(x_i) = \frac{1}{\exp(1 - x_i / \max_{j=1,\cdots,n} x_j)} \quad (30)$$

where $x_i$ is an independent variable, $\max_{j=1,\cdots,n} x_j$ represents the maximum value in a set $\{x_j | j = 1, \cdots, n\}$, and $n$ indicates the size of the set.

Based on empirical evidence (Chang et al., 2020), this operator will rank scheduled ground targets by the workpiece congestion in ascending order and the ground target with the larger workpiece congestion will be removed from the given solution in priority.

#### 4.3.2. "Repair" operators

All unscheduled ground targets are saved in a set denoted by $F$. The ground targets in $F$ but not in $B$ could be selected and inserted into a given solution to produce an offspring solution. RGHA is adopted in "Repair" operators to insert selected ground targets into the given solution. Four different "Repair" operators are defined as follows. The difference between them is their different guiding information for sorting the unscheduled ground targets.

**R-Repair**: This operator selects some unscheduled ground targets in $F$ but not in $B$ randomly and inserts them into the given solution.

**P-Repair**: The guiding information of this operator considers the priority of ground targets. This operator ranks all selected ground targets by the guidance information ascending and denotes the unscheduled ground target with higher priority will be considered in priority (Chang et al., 2020),

**L-Repair**: The guiding information of this operator considers the visible time window (VTW) of ground targets. For the ground target $gt_i$, the guiding information is defined as

$$GF\_L(gt_i) = gt_i.e - gt_i.s \qquad (31)$$

where $gt_i.e$ and $gt_i.s$ represents the begin time and end time of VTW of $gt_i$ respectively. This operator ranks all unscheduled ground targets by $GF\_L$ in ascending order and then the unscheduled ground target with shorter VTW will be considered in priority.

**C-Repair**: The guiding information of this operator is the same as that of **C-Destroy**. However, **C-Repair** operator inserts unscheduled ground targets with less workpiece congestion into the given solution in priority.

### 4.4. A variation of the box-method

The box-method (Hamacher et al., 2007) based on $\varepsilon$-constraint method is well known in solving multi-objective discrete optimization problems. It guides the evolution in the non-dominated region, and it is very useful in obtaining the Pareto frontier faster. On the other hand, the box-method does not decrease the diversity of solutions.

## 4.5. Adaptive layer and terminational criterion

Each operator has a score and a weight. The score depends on the performance of observation scheduling, and the weight is updated according to the score. Here, the four scores are defined as follows.

- $\sigma_1$   If the new solution dominates all current solutions
- $\sigma_2$   If the new solution dominates one of the current non-dominated solutions
- $\sigma_3$   If the new solution is located on the current Pareto frontier
- $\sigma_4$   If the new solution is dominated by one of the current non-dominated solutions

At the end of every iteration, the weights of each of the operators are updated using

$$\omega_i^\alpha = (1 - \lambda)\omega_i^\alpha + \lambda \frac{\pi_i^\alpha}{\sum_{j=1}^{|I_\alpha|} \pi_j^\alpha} \quad 1 \leq i \leq |I_\alpha| \tag{32}$$

where, $\alpha$ denotes the type of operator (Destroy or repair) and $|I_\alpha|$ represents the number of operators in the corresponding type. $\pi_i^\alpha$ and $\omega_i^\alpha$ denotes the score and weight of the i[th] operator and $\lambda \in [0,1]$ is a reaction factor that controls how sensitive the weights are to changes in the performance of operators. A value of 0 means that the weights remain unchanged, while a value of 1 implies that the historic performance has no impact and the weight only depends on the current score.

A roulette wheel mechanism is used to choose operators[6]. The utilization rate ($r_i^\alpha$) is calculated using the equation

$$r_i^\alpha = \frac{\omega_i^\alpha}{\sum_{j=1}^{|I_\alpha|} \omega_j^\alpha} \quad 1 \leq i \leq |I_\alpha| \tag{33}$$

The maximum number of iterations, denoted by $MaxIter$, is the only terminational criterion for ALNS+NSGA-II. The value of $MaxIter$ is given at the beginning of each evolution step.

## 5. Simulation experiments

In this section, we discuss results of extensive experimental analysis carried out using the proposed algorithms and operators. First, a simulation experiment is designed to analyze the effect of the *Envelop partition* proposed in (Wen-yuan et al., 2016). Then we will compare our algorithm, ALNS+NSGA-II, with HCBMDE proposed in (Wen-yuan et al., 2016). In the end, the evolution of all operators in ALNS+NSGA-II will be discussed and analyzed. All algorithms are coded in C#, using Visual Studio 2013, and the experiments were run on a laptop with intel(R) Core (TM) i7-8750H CPU @ 2.2GHz and 16 GB RAM.

---

[6] In the first iteration, all operators will be used in the equal possibility, then they will be chosen by the roulette wheel mechanism.

Good test cases are necessary for any experimental analysis of algorithms. However, there are no benchmark instances available for OSPFAS. Many researchers generated simulated instances based on real world data for related problems (Grasset-Bourdel, Verfaillie, & Flipo, 2011; Karapetyan, Mitrovic Minic, Malladi, & Punnen, 2015; X. L. Liu et al., 2017). We used the method of (X. L. Liu et al., 2017) for generating our test instances. In particular, we designed two types simulated instances, Chinese area distribution (CD) and Worldwide distribution (WD). Ten scenarios are generated in CD and the number of ground targets contained in these scenarios varies from 50 to 500, with an increment step size of 50. The center of all ground targets is randomly generated with a uniform distribution in the Chinese area (3°N-53°N and 74°E-133°E). For the class WD, the center of all ground targets is uniformly distributed worldwide and the number of ground targets varies from 100 to 1000, with an increment step size of 100. The number of vertices of every ground target belongs to the interval [3, 6] and selected randomly, the shape of every ground target is a convex polygon and the area of them is randomly generated and belongs to the interval [40, 2500], with the unit as $km^2$. Further, the priority of all ground targets is uniformly generated respectively from [1, 10].

Table 1 The parameters of the sensor

| $\theta$ | $\gamma$ | $\pi$ | $\psi$ | $d_0$ |
|---|---|---|---|---|
| 1.72° | 45° | 45° | 90° | 5s |

Since there are no in-orbit AI-AEOS in China and the orbit is not critical, we adopt the orbit parameters of Geofen-2 for building our simulation instances, and Geofen-2 choosing from the satellite database in the AGI Systems Tool Kit (STK) 11.2. We set the basic attributes of the sensor installed in Geofen-2 as shown in Table 1 and the scheduling horizon was 24 hours, from 2020/08/01 00:00:00 to 2020/08/02 00:00:00, to calculate the visible time windows (VTWs) for all ground targets. The whole data of satellite orbits and VTWs of all ground targets can be found in our data files[7] (available on request from the first author). Some general parameter settings for our algorithm ALNS+NSGA-II are shown in the Table 2.

---

[7] The time of all data is cumulative seconds based on 2020/08/01 00:00:00

Table 2 General parameters

| Parameter | Meaning | Value |
|---|---|---|
| $NS$ | The population size of all solutions/preserving solution | 100 |
| $NA$ | The population size of archive solutions | 100 |
| $MaxIter$ | The maximum number of iterations | 200 |
| $RS$ | The probability for selecting a ground target from $AGT$ | 0.2 |
| $TR$ | The rate of the size of taboo bank to the total ground targets | 0.2 |
| $\sigma_1$ | If the new solution dominates all current solutions; | 30 |
| $\sigma_2$ | If the new solution dominates one of the current non-dominated solutions; | 20 |
| $\sigma_3$ | If the new solution on the current Pareto frontier; | 10 |
| $\sigma_4$ | If the new solution is dominated by one of the current non-dominated solutions. | 0 |
| $\lambda$ | The value of the reaction factor to control update the weight of operators | 0.5 |

## 5.1. The effect of envelop partition

Our first set of experiments were focused on the partition sets. To illustrate visually the effect of the four partition sets, *ATO partition*, *NATO partition*, *Complete partition*, and *Envelope partition* on our algorithm ALNS + NSGA-II, we choose four simulation instances as examples, CD-50, CD-150, WD-100 and WD-200, and plot the Pareto frontiers obtained by ALNS+NSGA-II under different partition sets in Figure 3. The dotted green line denotes the Pareto frontier obtained by ALNS+NSGA-II under Envelope partition, the dotted black line, the dotted blue line and the dotted red line represent the Pareto frontier obtained by ALNS+NSGA-II under *ATO partition*, *NATO partition* and *Complete partition*, respectively.

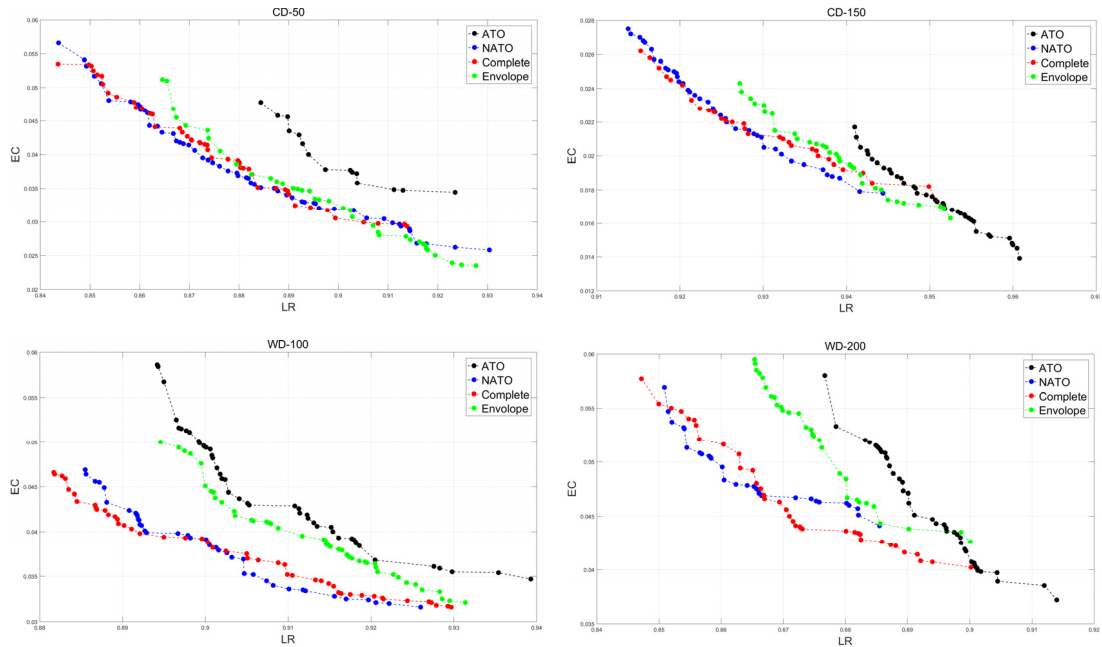

Figure 3 The Pareto frontiers of four instances under four partition sets

From the distribution of the Pareto frontiers under different partition sets, we can see that the elitist solutions obtained by ALNS+NSGA-II under *ATO partition* are always in the dominated regions of the other three partition sets. Which means the elitist solutions obtained by ALNS+NSGA-II under *ATO partition* are always worse regardless of the first objective (LR) or the second objective (EC). These results indicates that *NATO partition* is good for observing more ground targets with less energy.

Table 3 The values of indicators obtained by ALNS+NSGA-II under different partition sets

| Instances (NO. M) | Envelope partition | | | | ATO partition | | | | NATO partition | | | | Complete partition | | | |
|---|---|---|---|---|---|---|---|---|---|---|---|---|---|---|---|---|
| | $h$ | $t_p$ | $t_s$ | $t_w$ | $h$ | $t_p$ | $t_s$ | $t_w$ | $h$ | $t_p$ | $t_s$ | $t_w$ | $h$ | $t_p$ | $t_s$ | $t_w$ |
| CD-50 | 141.5 | 33 | 20 | 53 | 113.8 | 16 | 17 | 33 | 151.3 | 165 | 25 | 190 | 151.1 | 189 | 21 | 210 |
| CD-100 | 99.6 | 77 | 38 | 115 | 73.9 | 50 | 17 | 67 | 105.6 | 387 | 47 | 434 | 107.2 | 421 | 47 | 468 |
| CD-150 | 78.5 | 107 | 59 | 166 | 58.1 | 55 | 35 | 90 | 84.5 | 497 | 62 | 559 | 83.1 | 542 | 63 | 605 |
| CD-200 | 64.2 | 126 | 70 | 196 | 46.1 | 61 | 52 | 113 | 68.5 | 626 | 73 | 699 | 70.6 | 677 | 78 | 755 |
| CD-250 | 67.4 | 183 | 83 | 266 | 49.9 | 81 | 64 | 145 | 71.2 | 913 | 90 | 1003 | 71.1 | 978 | 91 | 1069 |
| CD-300 | 53.3 | 211 | 94 | 305 | 39.6 | 100 | 75 | 175 | 59.3 | 1098 | 126 | 1224 | 62.1 | 1198 | 121 | 1319 |
| CD-350 | 40.6 | 265 | 116 | 381 | 29.8 | 131 | 87 | 218 | 45 | 1400 | 142 | 1542 | 47.1 | 1527 | 146 | 1673 |
| CD-400 | 41.4 | 262 | 127 | 389 | 32.1 | 110 | 111 | 221 | 46.7 | 1431 | 152 | 1583 | 47.3 | 1561 | 143 | 1704 |
| CD-450 | 38.1 | 356 | 142 | 498 | 30.7 | 168 | 115 | 283 | 46.9 | 1890 | 169 | 2059 | 47.1 | 2086 | 182 | 2268 |
| CD-500 | 34.1 | 399 | 169 | 568 | 25.9 | 185 | 135 | 320 | 38.7 | 2095 | 193 | 2288 | 39.5 | 2319 | 193 | 2512 |
| WD-100 | 104.7 | 66 | 44 | 110 | 101.8 | 28 | 45 | 73 | 110.7 | 345 | 50 | 395 | 114.3 | 372 | 52 | 424 |
| WD-200 | 128.7 | 152 | 101 | 253 | 118.4 | 107 | 45 | 152 | 142.4 | 832 | 105 | 937 | 146.3 | 889 | 117 | 1006 |
| WD-300 | 103.1 | 210 | 147 | 357 | 99.9 | 150 | 89 | 239 | 114.8 | 1250 | 184 | 1434 | 115.3 | 1328 | 174 | 1502 |
| WD-400 | 108.2 | 307 | 205 | 512 | 102 | 192 | 149 | 341 | 116.2 | 1734 | 225 | 1959 | 118.8 | 1862 | 232 | 2094 |
| WD-500 | 98.3 | 526 | 332 | 858 | 89.1 | 246 | 215 | 461 | 114.4 | 2533 | 330 | 2863 | 116.4 | 2935 | 366 | 3301 |
| WD-600 | 118.3 | 534 | 339 | 873 | 93.5 | 369 | 283 | 652 | 123.6 | 3094 | 422 | 3516 | 126.6 | 3240 | 400 | 3640 |
| WD-700 | 96.8 | 643 | 505 | 1148 | 79.5 | 283 | 376 | 659 | 99.7 | 3410 | 546 | 3956 | 105.1 | 4316 | 487 | 4803 |
| WD-800 | 86.8 | 698 | 496 | 1194 | 70.2 | 332 | 545 | 877 | 91.4 | 4400 | 615 | 5015 | 96.5 | 4452 | 566 | 5018 |
| WD-900 | 89.4 | 864 | 585 | 1449 | 71.1 | 415 | 609 | 1024 | 95.4 | 6455 | 707 | 7162 | 98.5 | 6295 | 724 | 7019 |
| WD-1000 | 91.3 | 967 | 679 | 1646 | 73.6 | 414 | 676 | 1090 | 99.2 | 7596 | 871 | 8467 | 101.6 | 8712 | 939 | 9651 |

The location of the Pareto frontiers of *NATO partition* and *Complete partition* are near for these four simulation instances, and the elitist solutions obtained by ALNS+NSGA-II under them were found to be always better than those under the other two partition sets: *Envelope partition* and *ATO partition*. This suggests that the approach proposed in (Yang et al., 2018) where the *Envelop partition* is considered could produce inferior solution.

However, by considering only the quality of elitist solutions, we cannot simply confirm that the *Envelope partition* is a bad choice for MOSP. The running time of an algorithm is another important indicator, especially for problems in engineering practice (X. L. Liu et al., 2017). Thus, using all of the simulation instances, we will consider the running time and Hypervolume (HV) of the elitist solutions as other two performance indicators to analyze the efficiency of ALNS+NSGA-II under these four partition sets and it turns out that the *Envelope partition* became a better choice. Hypervolume (HV) is calculated by an algorithm, *hypervolume by slicing objectives* (HSO) (Bradstreet, While, & Barone, 2008; Durillo & Nebro, 2011). Set {1,1,1} as the reference point, the finial values of HV ($h$) after 200 iterations obtained by ALNS+NSGA-II under the four partition sets for all of the simulation instances are shown in Table 3. In addition, as mentioned above, because of the normalization, the values of the objectives are very small, so we multiply all values of $h$ by 1000. On the other hand, we would like to separate the running time of observation strips partition ($t_p$) from the whole running time ($t_w$) to show the effect of different partition sets. Here, $t_p$, $t_s$ and $t_w$ represent the partition running time, the scheduling running time and the whole running time respectively, with the unit chosen as seconds of CPU time.

The distribution of the Pareto frontier obtained by ALNS+NSGA-II under the four partition sets, the values of HV obtained by ALNS+NSGA-II under *ATO partition* are significantly smaller than that under the other three partition sets, which reinforces that it is important to consider NATO observation strips for MOSP.

The partition running time ($t_p$) forms a major portion of the whole running time $t_w$ for ALNS+NSGA-II under these four partition sets. As mentioned above, the feasible values of DOS of NATO strips belong to the interval [0°, 360°], with an increment step of 30°, *NATO*

*partition* and *Complete partition* consume more time for partition[8]. So, considering all NATO observation strips is not very desirable from the point of view of computational time.

The Pareto frontier obtained by ALNS+NSGA-II under *Envelope partition* is consistently better than that obtained by ALNS+NSGA-II under *ATO partition*, and not much worse than that obtained by ALNS+NSGA-II under *NATO partition* and *Complete partition*. Further, the running time consumed by ALNS+NSGA-II under *Envelope partition* is similar to that consumed by ALNS+NSGA-II under *ATO partition*, and significantly smaller than that consumed by ALNS+NSGA-II under *NATO partition* and *Complete partition*. In other words, the *Envelope partition* balances the quality of elitist solutions / Hypervolume (HV) and the running time to solve MOSP. Therefore, we conclude that *Envelope partition* is a better alternative when considering the partition set for solving MOSP. Thus, hereafter we only consider *Envelope partition* in our subsequent experiments.

## 5.2. The efficiency of ALNS+NSGA-II

To compare the relative efficiency of ALNS+NSGA-II in comparison to other algorithmic strategies, we selected a multi-objective optimization algorithm, called *hybrid coding based multi-objective differential evolution algorithm* (HCBMDE), proposed in (Yang et al., 2018). HCBMDE combines the differential evolution algorithm (DE) with NSGA-II. Since evidence exists that a random search can be competitive to evolutionary approaches in multi-objective spaces (Corne & Knowles, 2007; Purshouse & Fleming, 2007; R. Wang, Purshouse, & Fleming, 2013), we also designed a crude random selection mechanism (RSM) in which the elitist solutions are preserved randomly, and combine RSM with ALNS (ALNS+RSM) to produce another control algorithm. Using the simulation instances from the Chinese area distribution (CD), we restarted the three algorithms 50 times, and then got the boxplot of HV obtained by them in Figure 4. The black boxes, blue boxes and green boxes denote the HV obtained by ALNS+NSGA-II, HCBMDE and ALNS+RSM respectively. We also plot the Pareto frontiers of five specific simulation instances, CD-100 to CD-500, with an increment step of 100.

The location of black boxes is always higher than the other two types for all simulation instances, which means the value of HV obtained by ALNS+NSGA-II is always larger than that obtained by the other two algorithms.

---

[8] It is possible to reduce the partition running time to be some extent adopting the parallel computation. And adopting source code of partition method can also reduce the partition running time.

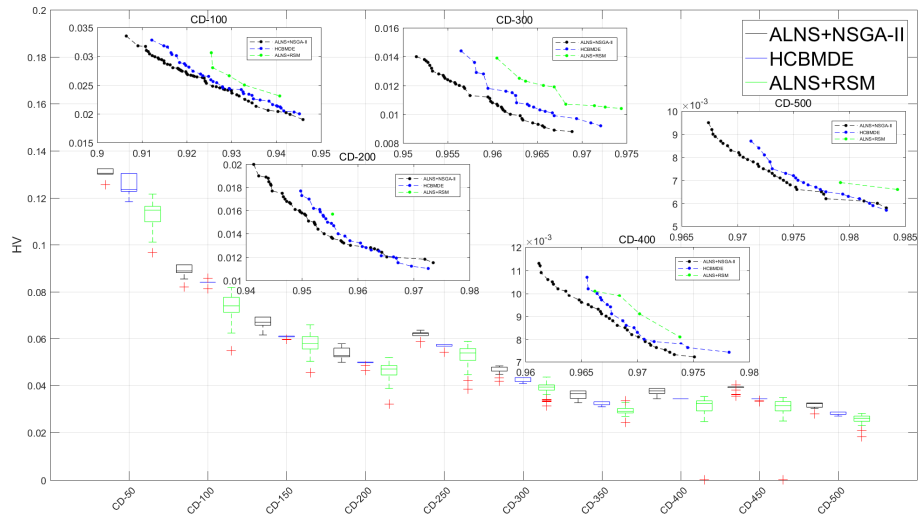

Figure 4 Boxplot: the value of HV obtained by ALNS+NSGA-II, HCBMDE and ALNS+RSM, and Dotted line: Pareto frontiers of five simulation instances obtained by these three algorithms

Since the length of black boxes and the blue boxes are shorter than that of the green boxes, both ALNS+NSGA-II and HCBMDE can produce good nondominated solutions consistently. On the other hand, fewer outliers of HV obtained by ALNS+NSGA-II and HCBMDE also indicate that both of these algorithms are viable choices for solving MOSP, and the performance of ALNS+NSGA-II however is much better.

For the five specific simulation instances, the location of Pareto frontiers also reflects that ALNS+NSGA-II is the best among algorithms compared. The location of Pareto frontiers obtained by ALNS+NSGA-II is always under that obtained by the other two algorithms. Further, Pareto frontiers obtained by ALNS+NSGA-II is also longer than that of the others. Note that the longer the Pareto frontier is, the more diverse the solutions are. Thus, we conclude based on various performance metrics ALNS+NSGA-II is a better alternative compared to the other algorithms used in our study.

Comparing with HCBMDE, ALNS+NSGA-II can search better nondominated solutions (bigger HV) and obtain much longer Pareto frontier, which shows that combining ALNS with NSGA-II is more suitable than combing DE with NSGA-II for solving MOSP. Also, in comparison to ALNS+RSM, ALNS+NSGA-II can achieve Pareto frontier in a stable manner.

## 5.3. Operators evolution

A simulation instance, CD-100 from the China Area distribution, is chosen to analyze the evolution of all proposed operators: "Destroy" operators and "Repair" operators. Let $\lambda$ belongs to the interval [0,1] with incremental steps 0.1, and restart ALNS+NSGA-II for 50 times using the instance CD-100. Then the final weights of all operator under different values of $\lambda$ are drawn by two boxplots in Figure 5 respectively, where the two boxplots correspond to two types operators. The y-axis denotes the values of final weight of operators, while the x-axis represents different values of $\lambda$. Note that the scale of the y-axis in two subplots are different.

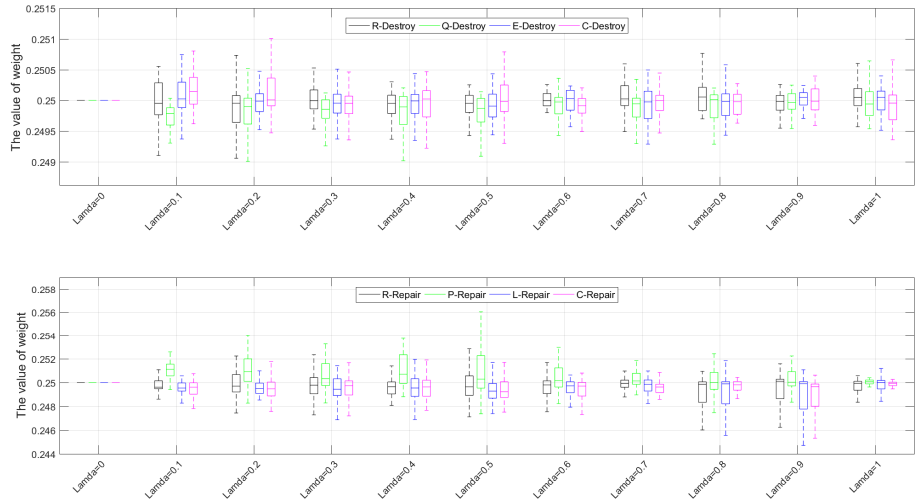

Figure 5 Boxplot: the final weight of all operators ("Destroy" operators and "Repair" operators) after restarting ALNS+NSGA-II for 50 times based on CD-100 under different values of $\lambda$

The mean values of the final weight for all operators ("Destroy" operators and "Repair" operators) under different values of $\lambda$ are almost located on the same horizontal line around 0.25. This phenomenon reflects that the efficiency of all proposed operators under different values of $\lambda$ are similar. As mentioned above, when $\lambda$ equals zero, the iteration weights of operators remain unchanged. In other words, there is no significant influence of $\lambda$, on the performance of the algorithm and we set $\lambda = 0.5$ for our experiments.

Note that the length of the box represents the deviation. Since the boxes of "Repair" operators are always longer than that of "Destroy" operators, the "Destroy" operators are more efficient and stable.

As mentioned above, "Destroy" operators and "Repair" operators should be used in pairs. When $\lambda$ equals 0.5, the mean of final weight of C-Destroy operator is slightly larger than others among "Destroy" operators, while the mean of final weight of P-Repair operator is slightly larger than others among "Repair" operators. So, adopting C-Destroy and P-Repair in pairs is more favorable for generating better solutions.

## Conclusion

In this paper, we study the multi-strip observation scheduling problem for an active-imaging agile earth observation satellite (MOSP). A bi-objective optimization model considering the loss rate of image quality (LR) and the energy consumption (EC), is developed to represent MOSP. Under a memetic computing/algorithm (MC/MA) framework, we developed an adaptive bi-objective memetic algorithm utilizing careful integration of an adaptive large neighborhood search algorithm (ALNS) and a nondominated sorting genetic algorithm II (NSGA-II). We also proposed a randomized greedy heuristic algorithm (RGHA) as the initialization algorithm. Two operators, "Destroy" and "Repair", are introduced to generate our adaptive large neighborhoods (ALNs). The efficiency of the proposed algorithms and operators is analyzed using systematic computational experiments on simulated data.

We first compared our ALNS+NSGA-II algorithm under different partition sets, *ATO partition*, *NATO partition*, *Complete partition* and *Envelope partition*, to evaluate the effect of these partition sets on the quality of elitist solutions produced and the impact on the running time of algorithm. The results disclosed that, on balance the *Envelope partition* is the most suitable partition set for solving MOSP. However, basing purely of solution quality, *NATO partition* and *Complete partition* are favored.

We also compared our algorithm with HCBMDE proposed in (Yang et al., 2018) and another control algorithm, ALNS+RSM, to analyze the relative efficiency of ALNS+NSGA-II. Our ALNS+NSGA-II algorithm turned out to be more robust and produced superior solution quality with longer Pareto frontiers.

Finally, the impact of our "Destroy" operators and "Repair" operators is analyzed in depth. The boxplots of the final weight of all operators show that the effect of all proposed operators is similar under different values of $\lambda$.

Our algorithm produced a novel solution framework for solving MOSP and it outperformed competing algorithms on various performance metrics, including solution quality, running time, and robustness. The effect of each components of the algorithm is examined independently. The results disclosed that the algorithm with the individual components when worked in unison produced superior outcomes.

**Acknowledgements:** This work was completed when Zhongxiang Chang visited Abraham P. Punnen at the Simon Fraser University, Canada under the CSC scholarship program of the Peoples Republic of China. The research of Abraham P. Punnen was partially supported by an NSERC discovery grant. The research of Zhongxiang Chang was also supported by the science and technology innovation Program of Hunan Province (2021RC2048).